\documentclass[runningheads]{llncs}

\usepackage{eccv}

\usepackage{eccvabbrv}

\usepackage{graphicx}
\usepackage{booktabs}

\usepackage[accsupp]{axessibility}  

\usepackage{array}
\newcolumntype{L}[1]{>{\raggedright\let\newline\\\arraybackslash\hspace{0pt}}m{#1}}
\newcolumntype{C}[1]{>{\centering\let\newline\\\arraybackslash\hspace{0pt}}m{#1}}
\newcolumntype{R}[1]{>{\raggedleft\let\newline\\\arraybackslash\hspace{0pt}}m{#1}}

\usepackage[pagebackref,breaklinks,colorlinks,citecolor=eccvblue]{hyperref}

\usepackage{orcidlink}

\usepackage{comment}
\usepackage{enumitem}
\usepackage{multirow}
\usepackage{wrapfig}
\usepackage{colortbl}
\usepackage[normalem]{ulem}

\usepackage{caption}
\usepackage{graphicx}
\usepackage{tabularx}
\usepackage{booktabs}
\usepackage{subcaption}
\usepackage{tcolorbox}

\setlist[itemize]{align=parleft,left=0pt}

\makeatletter
\def\@fnsymbol#1{\ensuremath{\ifcase#1\or *\or \dagger\or \ddagger\or
   \mathsection\or \mathparagraph\or \|\or **\or \dagger\dagger
   \or \ddagger\ddagger \else\@ctrerr\fi}}
\makeatother

\def\onedot{.\@\xspace}
\def\eg{\emph{e.g}\onedot} 
\def\ie{\emph{i.e}\onedot}

\newcommand{\Sref}[1]{Sec.~\ref{#1}}

\newcommand{\Fref}[1]{Fig.~\ref{#1}}
\newcommand{\Tref}[1]{Table~\ref{#1}}

\newcommand{\calS}{{\mathcal{S}}}

\newcommand{\be}{\begin{eqnarray}}
\newcommand{\ee}{\end{eqnarray}}
\newcommand{\bee}{\begin{eqnarray*}}
\newcommand{\eee}{\end{eqnarray*}}

\newcommand{\matrixb}{\left[ \begin{array}}
\newcommand{\matrixe}{\end{array} \right]}

\definecolor{green(ncs)}{rgb}{0.0, 0.62, 0.42}

\renewcommand{\paragraph}[1]{\noindent\textbf{#1.}\,\,}

\begin{document}

\title{BEAF: Observing BEfore-AFter Changes to Evaluate Hallucination in Vision-language Models} 

\titlerunning{BEAF Benchmark}

\author{Moon Ye-Bin\inst{1}\thanks{Authors contributed equally to this work.}\orcidlink{0000-0002-0390-6567} \ 
Nam Hyeon-Woo\inst{1\ast}\orcidlink{0000-0001-9543-3770} \ 
Wonseok Choi\inst{2}\orcidlink{0009-0008-2665-2388} \ 
Tae-Hyun Oh\inst{1,2,3}\orcidlink{0000-0003-0468-1571}}

\authorrunning{Ye-Bin et al.}

\institute{
${}^1$Dept. of EE, POSTECH, Korea \qquad ${}^2$Grad. School of AI, POSTECH, Korea\\
${}^3$Institute for Convergence Research and Education in Advanced Technology, Yonsei University, Korea.\\
\email{\{ybmoon, hyeonw.nam, wonseok.c, taehyun\}@postech.ac.kr}
}

\maketitle

\begin{abstract}

Vision language models (VLMs) perceive the world through a combination of a visual encoder and a large language model (LLM). The visual encoder, pre-trained on large-scale vision-text datasets, provides zero-shot generalization to visual data, and the LLM endows its high reasoning ability to VLMs. It leads VLMs to achieve high performance on wide benchmarks without fine-tuning, exhibiting zero or few-shot capability. However, recent studies show that VLMs are vulnerable to hallucination. This undesirable behavior degrades reliability and credibility, thereby making users unable to fully trust the output from VLMs. To enhance trustworthiness and better tackle the hallucination of VLMs, we curate a new evaluation dataset, called the BEfore-AFter hallucination dataset (BEAF), and introduce new metrics: True Understanding (TU), IGnorance (IG), StuBbornness (SB), and InDecision (ID). Unlike prior works that focus only on constructing questions and answers, the key idea of our benchmark is to manipulate visual scene information by image editing models and to design the metrics based on scene changes. This allows us to clearly assess whether VLMs correctly understand a given scene by observing the ability to perceive changes. We also visualize image-wise object relationship by virtue of our two-axis view: vision and text. Upon evaluating VLMs with our dataset, we observed that our metrics reveal different aspects of VLM hallucination that have not been reported before. Project page: \url{https://beafbench.github.io/}

\end{abstract}

\section{Introduction}\label{sec:intro}
Vision language models (VLMs)~\cite{achiam2023gpt4, chen2023shikra, instructblip, liu2023improvedllava, liu2023llava} have recently emerged by its multi-modal reasoning capability. It is indeed promising research toward making intelligent agents perceive the world. VLMs comprise a vision encoder as a visual perception module and a large language model (LLM) as a reasoning module. Recent LLMs~\cite{touvron2023llama, touvron2023llama2, vicuna2023} have exhibited powerful reasoning capabilities across various tasks, including commonsense reasoning~\cite{piqa2020, hyun2023smile} and math~\cite{gsm8k2021}, in zero-/few-shot ways. This success motivates researchers to integrate LLMs with vision encoders to open LLMs' eyes to see the world. Leveraging the reasoning capabilities of LLMs alongside visual perception enables VLMs to excel in various visual tasks, such as image captioning~\cite{flickr30k2014, nocaps2019} and visual question answering ~\cite{gqa2019, balanced_vqa_v2} without task-specific fine-tuning.

Despite this encouraging demonstration, VLMs are susceptible to the hallucination~\cite{li2023pope, hu2023ciem, gunjal2023mhaldetect, wang2023amber, zhai2023halle, zhou2023lure} wherein their outputs do not reflect facts presented in input images. For example, although a given input image does not present an object, VLMs often incorrectly answer for the presence of the object. This tendency is problematic, akin to spreading falsehoods in human communication. Hallucination prevents genuine communication, consequently diminishing the reliability of interactions between agents or humans. No matter how good VLM's answer is, if hallucination is included in the answer, it can undermine trust in the system's reliability and credibility. Therefore, it is crucial to address hallucinations to improve the system's responsibility and trustworthiness.

\begin{figure}[t]
    \centering
    \includegraphics[width=0.96\linewidth]{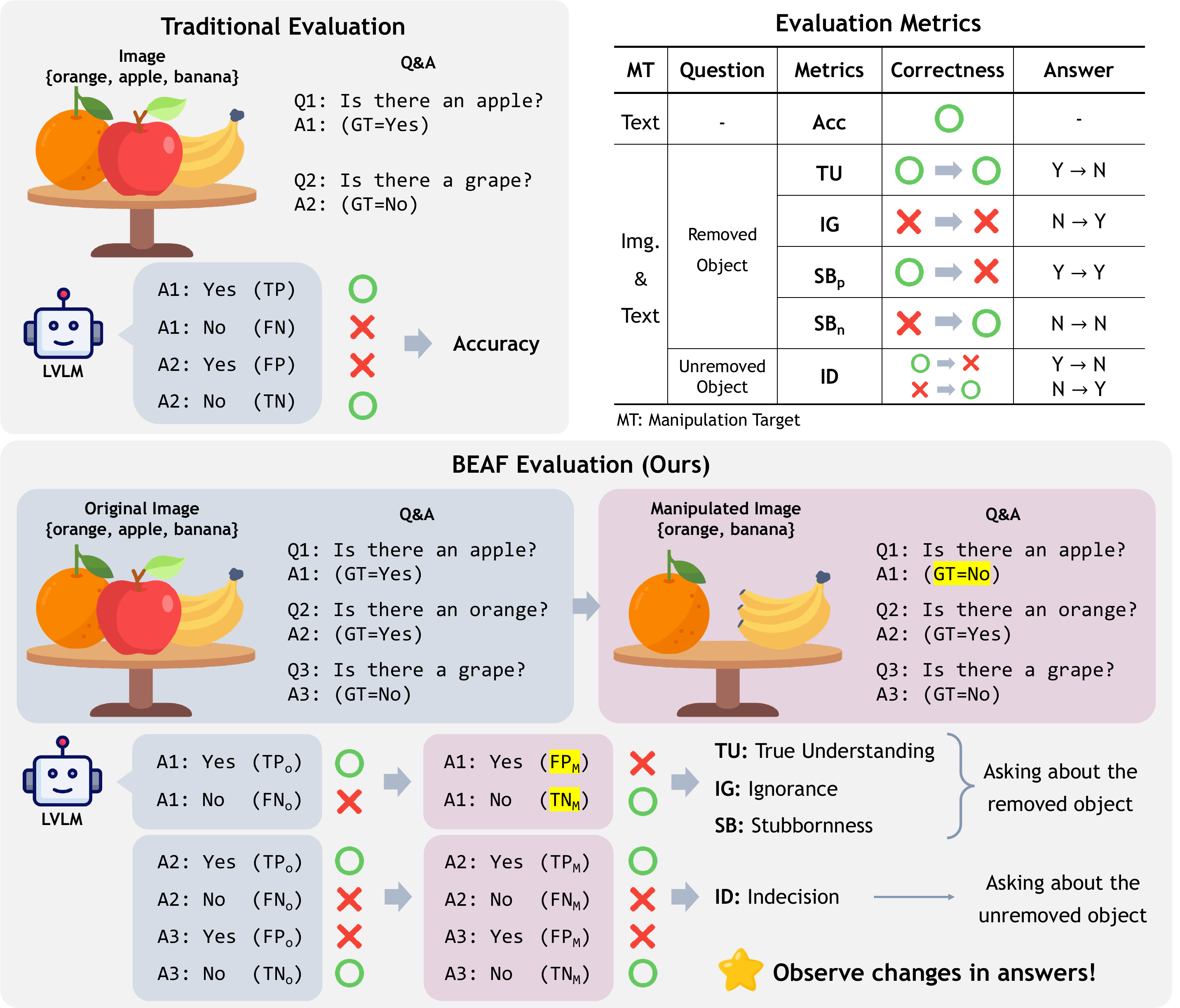}
    \caption{\textbf{Illustration of BEfore-AFter (BEAF) benchmark.} 
    We present a comparison between traditional evaluation benchmarks and our BEAF benchmark for assessing the hallucination behavior in VLMs.
    Traditional evaluation methods solely manipulate questions based on the existence of an object and measure accuracy or F1 score.
    In contrast, our BEAF benchmark not only constructs questions but also manipulates images and tracks changes in answers as the images undergo manipulation.
    The BEAF benchmark introduces novel metrics including True Understanding (TU), IGnorance (IG), StuBbornness (SB), and InDecision (ID), which consider the changes in answers.
    }
    \label{fig:main}
\end{figure}

Understanding the impact of hallucinations on model performance requires the development of an assessment framework, which requires a deep exploration of the underlying causes.
The prior works~\cite{hu2023ciem, zhou2023lure, zhao2023enhancing} have been proposed to unveil the causation of hallucinations in terms of data and models by reporting their performance on existing hallucination benchmarks. 
As a widely used hallucination benchmark, POPE~\cite{li2023pope} follows the question-and-answer (QnA) format.
For the convenience of measuring, POPE uses a discriminative question, \eg, ``\texttt{Is there \{object\} in the image?}'', and gets a YES or NO answer.
The following hallucination benchmarks~\cite{li2023pope, hu2023ciem, wang2023amber} commonly leverage LLMs or human annotators to generate question prompts with available scene information such as bounding boxes and captions.
This text-axis evaluation may assess certain aspects of the hallucination performance well, but it is hard to disentangle sources of hallucination, particularly in VLMs that handle both visual and text modalities.
Thus, to thoroughly analyze and understand the behavior of VLMs, it is both natural and necessary to consider both modalities.

In this work, we propose a BEfore-AFter (BEAF) hallucination evaluation benchmark.
This benchmark comprises a dataset enhanced along the vision-axis and metrics that are aware of changes.
We manipulate scenes by selectively removing objects on the vision-axis while augmenting QnA pairs along the text-axis. 
We can observe how the models' answers change as the image is manipulated for the same question.
For example, as illustrated in \Fref{fig:main}, if we remove an apple from the given image and then ask ``\texttt{Is there an apple?}, the answer from the model should change from ``\texttt{Yes}'' to ``\texttt{No}'', between the original and modified images, if it truly relies on what it sees.
Therefore, for a more thorough evaluation and analysis, we need change-aware metrics beyond just using standard accuracy~\cite{li2023pope, hu2023ciem, wang2023amber}.

We propose four new metrics: True Understanding (TU), IGnorance (IG), StuBbornness (SB), and InDecision (ID), which are enabled by our vision and text augmentation.
TU measures whether the model answers correctly before and after the manipulation.
IG evaluates whether the models perceive the images consistently both before and after manipulation.
SB is related to the phenomenon of giving the same answer instead of correctly recognizing the image.
ID measures cases where the answer changes even though the target object does not change or does not exist in the image.
The first three metrics address questions regarding target objects that have been removed, while the last one belongs to questions about other objects not related to the targets.

Based on our BEAF dataset and change-aware metrics, we can evaluate and understand the models, even the parts we did not know through the previous methods.
The evaluation results imply that answers assumed as non-hallucination from the existing hallucination benchmarks could be a hallucination.
We also analyze the influence of each object in the scene based on the variation along the vision and text axes.
In addition, we include a discussion on open-generation answers and CLIPScore results.
Our key contributions are as follows:
\begin{itemize}
    \item We propose a benchmark, called BEfore-AFter (BEAF), to evaluate the hallucination of large vision-language models.
    Our benchmark comprises a dataset that considers both text- and vision-axes, and change-aware metrics, which allow us to perform granular evaluation.
    \item We demonstrate that previously reported outcomes deemed non-hallucinatory in the prior text-axis-only evaluation benchmarks may be hallucinations.
    \item We also visualize the impact of relationships between objects within hallucinatory images based on the BEAF results.
\end{itemize}
\section{Related Work}\label{sec:rw}
\paragraph{Vision Language Models (VLMs)}
VLMs~\cite{instructblip, liu2023llava, liu2023improvedllava, gao2023llamaadapterv2, zhang2023llamaadapter, chen2023shikra, peng2023kosmos, achiam2023gpt4, zhu2023minigpt4, chen2023minigpt4v2} have demonstrated their impressive image describing and reasoning ability based on powerful large language models (LLMs)~\cite{vicuna2023, touvron2023llama, touvron2023llama2, taori2023alpaca}.
Generally, VLMs consist of three parts: vision encoder, LLM, and vision-language alignment module. 
The vision information extracted from the vision encoder is translated to the LLM-understandable token by passing through the alignment module. LLM takes a prompt text and visual tokens as input and outputs response text tokens according to a given image and prompt.
Their training process is divided into multiple steps depending on which parts of VLMs are trained.

\paragraph{Hallucination in VLMs}
Hallucination~\cite{liu2024survey, zhai2023halle, gunjal2023mhaldetect, li2023pope, wang2023evaluation, wang2023amber} in VLMs refers to the case where the generated text answer does not reflect the true contents of the provided images but rather relies on the internal knowledge of the models. This behavior might cause harmful consequences in high-risk applications such as the medical domain or auto-driving scenarios. This undesirable behavior could be caused by various factors, $\textit{e.g.}$, data imbalance, image resolution, and model capacity. The assessment of hallucination is also related to the true understanding of VLMs. If VLMs respond to the given input correctly rather than relying on innate knowledge, VLMs are considered to have a perfect understanding of the perceived world. Numerous methodologies~\cite{gunjal2023mhaldetect, zhou2023lure, liu2023lrvinstruction} have been developed to tackle the issue of hallucination in VLMs by addressing data biases~\cite{hu2023ciem, liu2023lrvinstruction}, optimization techniques~\cite{sun2023llavarlhf, leng2023vcd, gunjal2023mhaldetect}, and post-processing methodologies~\cite{zhou2023lure, yin2023woodpecker, deng2024seeing}. 

\paragraph{Evaluating Hallucination in VLMs}
Evaluating hallucination performance is essential to guide the research of VLM. Previous studies have introduced hallucination benchmarks to understand and assess these undesirable phenomena of VLMs. Motivated by traditional visual and question benchmarks, existing hallucination benchmarks~\cite{li2023pope,hu2023ciem,wang2023amber} follow this VQA style that provides a simple QnA evaluation pipeline and intuitive metrics. POPE~\cite{li2023pope} and CIEM~\cite{hu2023ciem} curate images and QnA pairs with metrics such as accuracy, precision, and recall, where they use a specific question type inducing multi-choice or yes/no answers, called discriminative type. AMBER~\cite{wang2023amber} includes an additional question type that induces open-ended caption generation as answers, called generative type. For such open-generation assessment, they extract the objects from the output text answers and compare them with the ground-truth objects in the image. These prior studies focus on manipulating texts of questions and answers only, which we call text-axis evaluation. However, if some objects always co-occur with another object, we cannot judge whether VLMs truly understand scene information accurately. Also, VLMs take multi-modal inputs, both images and text prompts; thus, evaluation considering only the text-axis would be insufficient as a hallucination assessment of VLMs. In this work, we manipulate the scene information and questions together, \ie, vision-/text-axes; it enables the fine-grained analysis of hallucination. We call it vision-/text-axes evaluation.

\section{BEAF Evaluation Benchmark for Hallucination}\label{sec:2deval}

We explore another axis, \ie, vision-axis, to make up for the deficiencies in the text-axis evaluation.
Our key idea is to manipulate a visual scene by removing and inpainting objects within an image.
This allows us to focus on changes and to check whether the VLMs are correctly aware of it. We describe how to curate our dataset in \Sref{subsec:pipe}, and then elaborate our new metrics in \Sref{subsec:metric}

\begin{table}[t]
    \centering
    \caption{\textbf{Data statistics.} 
    Our BEAF dataset contains 26K image-question pairs, consisting of the original and manipulated ones.
    On average, an original image is associated with 3.45 manipulated images and 11.72 questions.
    }
    \label{tab:data_statistics}
    \resizebox{0.8\textwidth}{!}{ 
    \begin{tabular}{@{\ \ }l@{\quad}C{20mm}C{20mm}C{20mm}C{20mm}}
        \toprule
                    & \textbf{Image} & \textbf{Img-Q Pair} & \textbf{Yes}   & \textbf{No}\\
        \midrule
        Original    & 500   & 5,653      & 2,026 & 3,627  \\
        Manipulated & 1,727 & 20,465     & 6,267 & 14,198 \\
        \midrule
        Total       & 2,227 & 26,118     & 8,293 & 17,825 \\
        \bottomrule
    \end{tabular}
    }
\end{table}

\begin{figure}[t]
    \centering
    \includegraphics[width=1.0\linewidth]{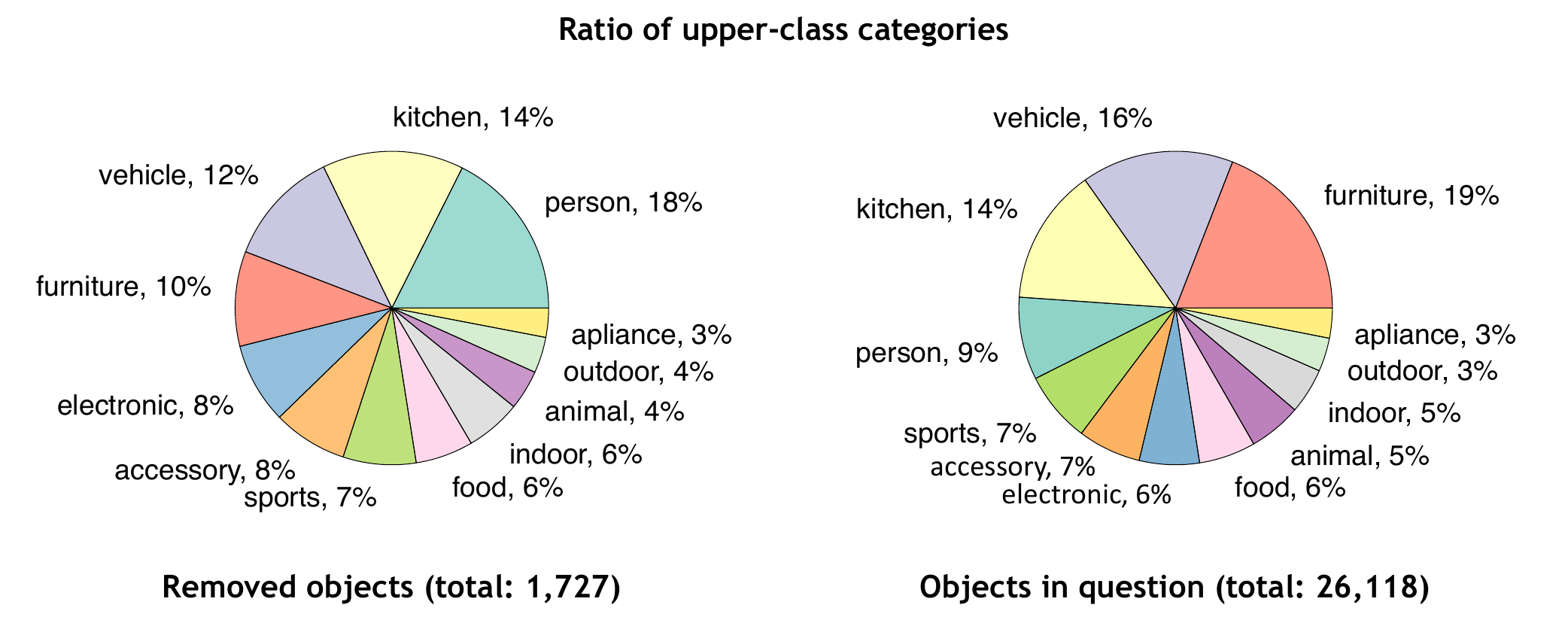}
    \caption{\textbf{Ratio of removed objects from image and object in question.} 
    For convenience, we report the ratio of upper-class categories instead of the object class itself.
    The total number of removed objects is the same as the number of manipulated images, and one object in question is the same as the number of image-question pairs.
    }
    \label{fig:statistics}
\end{figure}

\subsection{Evaluation Dataset}\label{subsec:pipe}

\paragraph{Dataset Overview}
\Tref{tab:data_statistics} and \Fref{fig:statistics} summarize our data statistics. The dataset consists of images and their associated QnA. The image type is divided into original and manipulated.
Five hundred original images are sampled from the validation set of the MS-COCO dataset~\cite{lin2014microsoft}, which shares the same split with POPE~\cite{li2023pope}, and their associated 1,727 manipulated images are generated by removing objects from the original images. On average, 3.45 manipulated images are created from the original images.
The main question format is ``\texttt{Is there \{object\} in this image?}'' and \texttt{\{object\}} can be a positive object within the image or a negative object not in the image;
the positive object is one of the objects in the image, and the negative object is the one not in the image.
We adopt the positive and negative objects from the POPE for the original image and adapt the question for the manipulated image.
If we remove the positive object from the original image to make the manipulated one, the positive object becomes the negative object in the manipulated one.
The total dataset comprised 2,227 images and 26,118 image-question pairs.
The ``\texttt{Yes}'' and ``\texttt{No}'' ground truth answers are 8,293 (31.75\%) and 17,825 (68.25\%), respectively.

\begin{figure}[t]
    \centering
    \includegraphics[width=1.0\linewidth]{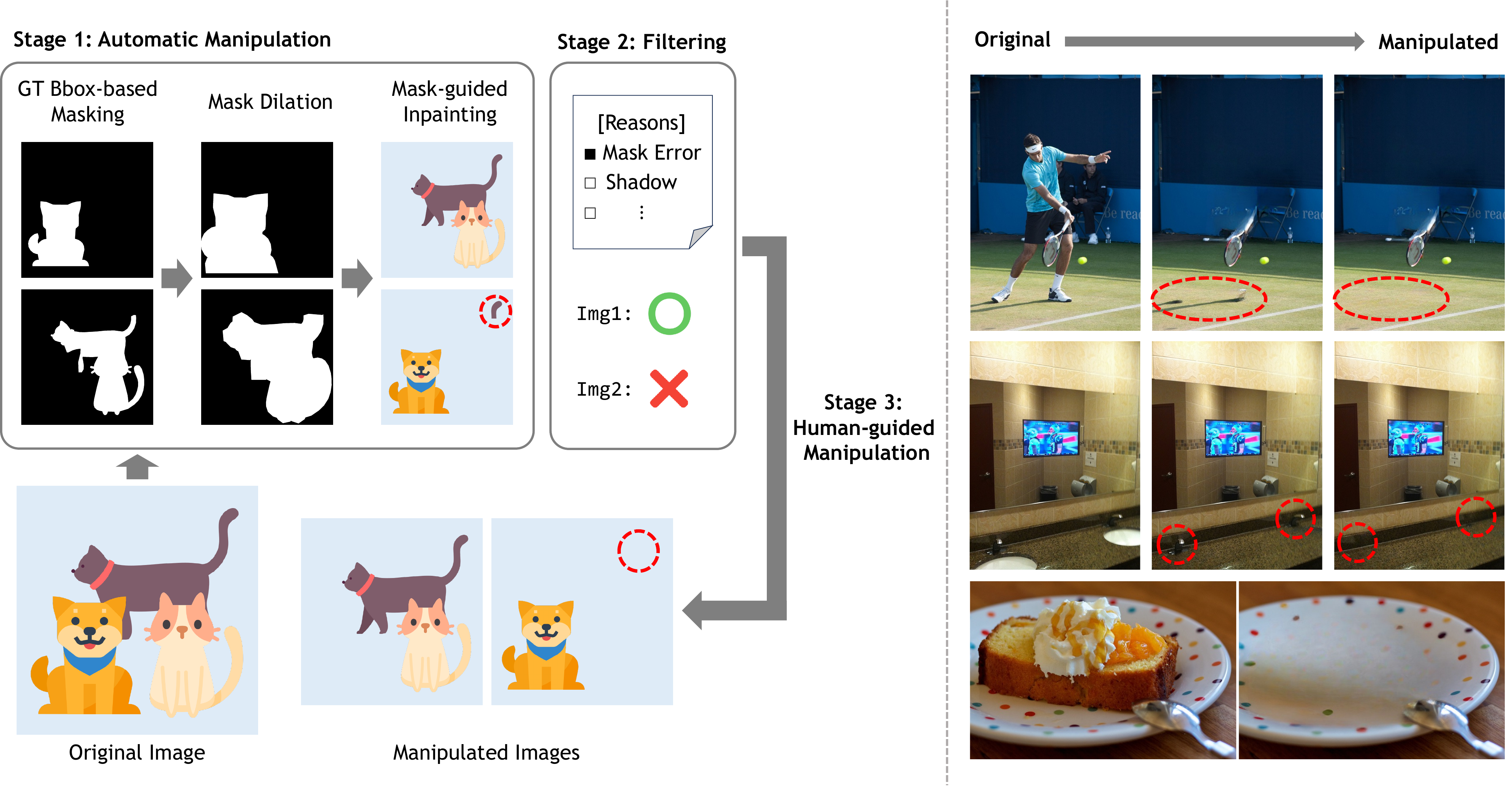}
    \caption{\textbf{Image manipulation pipeline.} 
    We illustrate the image manipulation pipeline comprised of three stages.
    In \textbf{Stage 1}, we automatically remove target objects sharing the same semantic class from the given images.
    \textbf{Stage 2} is to filter the automatically manipulated results based on the predefined rules, such as mask errors, remained shadows, and low-quality outcomes. 
    Undesirable
    manipulations are either corrected or discarded during this stage.
    Finally, in \textbf{Stage 3}, human annotators 
    engage in human-guided manipulation of the filtered images to achieve high-quality results.
    } 
    \label{fig:pipeline}
\end{figure}

\paragraph{Image Dataset Construction Pipeline}
We manipulate the original images by removing objects within the image. To collect the manipulated images, we pass through a three-stage pipeline (see \Fref{fig:pipeline}): the first stage involves automatic manipulation, followed by a filtering stage, and finally, a human-guided manipulation stage. We offer a detailed explanation of each stage as follows.

\textit{1) First Stage: Automatic Manipulation.} 
For automatic object removal, we use an off-the-shelf inpainting model, LaMa~\cite{suvorov2021lama}, with mask guidance.
While the initial option is to use the ground truth masks provided within the original dataset, we opt not to use them because of the low quality ground truth masks in MS-COCO. 
For example, polygon masks often cannot cover all areas of the corresponding object and can be propagated to the inpainting quality.
Therefore, we employ a pre-trained mask extractor called SAM~\cite{kirillov2023sam}, utilizing bounding box annotations. 
Subsequently, we dilate the extracted mask to have a margin for more effective object removal, and remove the object with the inpainting model given the dilated mask guidance.
We empirically choose to use LaMa over diffusion models, \eg, \cite{rombach2022high,zhuang2023task}, as we find that this specialized model performs better in the context of object removal.

\textit{2) Second Stage: Filtering.}
Although we remove objects from the original images, achieving consistent quality in the outcome can be influenced by a myriad of factors. 
These factors include, but are not limited to, the size of the objects being removed and the complexity of the backgrounds they inhabit. 
Additionally, the presence of shadows remains unaddressed, as they are not effectively masked using SAM; the ground truth masks also do not capture the shadow. 
As shown on the right side of \Fref{fig:pipeline}, it displays undesirable images due to the presence of shadow or a faucet. 
The human annotators classified the results obtained from the previous stage into usable or unusable, where only the usable results are further manipulated.
The detailed filtering rules can be found in the supplementary material.
We directly use the usable results and hand over the results that need further manipulation to the next stage.

\textit{3) Third Stage: Human-guided Manipulation.} 
The third stage of our workflow relies on human annotators to intervene and refine the images obtained from the second stage. 
As mentioned above, these filtered images often exhibit undesirable artifacts such as ghost shadows, afterimages, and fragmented objects. 
Even such a small artifact may provide strong cues indicating the past presence of objects that both humans and machines can deduce, $\textit{e.g.}$, a cat's tail suggests the past presence of the cat before the object removal. 
Thus, human annotators are asked to meticulously manipulate the images, either correcting or enhancing them. 
Annotators effectively refine visual data, aligning it more closely with real natural images in human perception.

As shown in \Fref{fig:pipeline}, our annotation process results in images that are free of the aforementioned artifacts.
This manual intervention is crucial, as it ensures the fidelity and reliability of our benchmark dataset.

\subsection{Evaluation Metrics}\label{subsec:metric}
The prior studies~\cite{li2023pope, hu2023ciem, wang2023amber} suggest using traditional metrics such as accuracy, precision, recall, and F1 score, which we can also use the metrics in our dataset.
In addition, we propose four new metrics: True Understanding (TU), IGnorance (IG), StuBbornness (SB), and InDecision (ID), for more detailed evaluation by exploiting the distinctive configuration of before-/after-changes in our dataset.

Let $\calS = \{(I_o, I_m, Q, A_o, A_m, R)_i\}_{i=1}^{|\calS|}$ denote our dataset, where each tuple consists of an original image $I_o$, a version of its manipulated image $I_m$, their associated question $Q$, their respective corresponding answers $A_o$ and $A_m$, and a True/False flag $R$ that indicates whether the question is associated with the object being removed in the manipulated image $I_m$. 
For compactness of our metric definitions, we define $\textsf{Filter}(\cdot)$ to extract the subset of $\calS$ satisfying input conditions as:
\begin{align*}
    \mathsf{Fil}&\mathsf{ter} ( b_o, b_m, b_r) = \{ \mathtt{t} | 
    \mathsf{IsCorrect}(A_o) = b_o , \mathsf{IsCorrect}(A_m) = b_m , R=b_r, 
    \mathtt{t} \in \calS
    \}
\end{align*}
where the tuple $t=(I_o, I_m, Q, A_o, A_m, R)$. 
For example, $\mathsf{Filter}(\mathtt{True, False, True)}$ is the subset of $\calS$ of which the element satisfy
the conditions that are 1) the answers corresponding to respective original and manipulated images are correct and wrong, and 2) the question is the one related to the object removed in the manipulated images.

\paragraph{True Understanding}
True Understanding (TU) measures whether the models truly understand the scene.
It evaluates if the model can accurately answer to the question about the removed object.
As shown in \Fref{fig:main}, if the apple is removed, a model that truly understands the change should switch its answer from ``\texttt{Yes}'' to ``\texttt{No}'' to the question ``\texttt{Is there an apple?}''. 
We define the TU metric as follows:
\begin{align}
    \text{TU} = 
    \tfrac{|\mathsf{Filter} (\mathtt{True}, \mathtt{True}, \mathtt{True})|}{|\mathsf{Filter} \textsf{(} \mathtt{True} \vee \mathtt{False},\, \mathtt{True} \vee \mathtt{False},\, \mathtt{True}) |} \cdot 100.
\end{align}
Compared to the traditional accuracy, our metric specifically considers the correctness of answers before and after manipulation. 
Therefore, it provides a conditional measure of accuracy, taking into account the model's response to changes within the scene.
A higher value of this metric indicates a more accurate understanding of the scene by the model, as it demonstrates the model's ability to adapt its answers appropriately to the scene changes.

\paragraph{Ignorance}
The second metric is IGnorance (IG), which measures the extent to which models lack knowledge about specific scene information. 
This metric quantifies ignorance by identifying instances where models provide incorrect answers to the question about the removed objects.
As shown in \Fref{fig:main}, ignorance is evident if models incorrectly answer ``\texttt{No}'' to the presence of an apple before its removal and incorrectly answer ``\texttt{Yes}'' after the apple is removed, indicating a failure to recognize it. 
IG is defined as follows:
\begin{align}
    \text{IG} = \tfrac{|\mathsf{Filter} (\mathtt{False}, \mathtt{False}, \mathtt{True})|}{|\mathsf{Filter} (\mathtt{True} \vee \mathtt{False}, \mathtt{True} \vee \mathtt{False}, \mathtt{True}) |} \cdot 100.
\end{align}
The lower the metric, the less ignorant the model is.

\paragraph{Stubbornness}
What we observe is that VLMs tend to output the same answer repeatedly.
It means that VLMs are biased toward the specific answer rather than arbitrary one when they do not know. 
This repetitive behavior is undesirable; thus, we introduce the metric of StuBbornness (SB), which measures the extent to which models adhere to their initial answers.
Furthermore, we categorize SB into SB$_{p}$ and SB$_{n}$ where the subscripts of $p$ and $n$ correspond to the consistent positive (``\texttt{Yes}'') and negative (``\texttt{No}'') answers, respectively.
We define them as follows:
\begin{align}
    \hspace{-2mm}\text{SB$_p$}  = \tfrac{ 100 \cdot|\mathsf{Filter} (\mathtt{True}, \mathtt{False}, \mathtt{True})|}{|\mathsf{Filter} (\mathtt{True} \vee \mathtt{False}, \mathtt{True} \vee \mathtt{False}, \mathtt{True}) |},\,\,\, 
    \text{SB$_n$}  = \tfrac{100 \cdot|\mathsf{Filter} (\mathtt{False}, \mathtt{True}, \mathtt{True})|}{|\mathsf{Filter} (\mathtt{True} \vee \mathtt{False}, \mathtt{True} \vee \mathtt{False}, \mathtt{True}) |}.
\end{align}
SB is the summation of SB$_{p}$ and SB$_{n}$ and can be computed from $100 - \text{TU} - \text{IG}$. 
The lower the metric, the less ignorant the model is.

\paragraph{Indecision}
Unlike the previous three metrics that focus on the removed objects in the manipulated images, InDecision (ID) focuses on the answers to the questions that are not relevant to the removed objects.
These answers should not be changed even after the manipulation.
Nevertheless, models, often change their answers about unmanipulated objects, which would imply that the model predicts a random answer without grounding it on the image.
We track these behaviors with the ID metric defined as follows:
\begin{align}
    \text{ID} = \tfrac{|\mathsf{Filter} (\mathtt{True}, \mathtt{False}, \mathtt{False})| + |\mathsf{Filter} (\mathtt{False}, \mathtt{True}, \mathtt{False})|}{| \mathsf{Filter}(\mathtt{True} \vee \mathtt{False}, \mathtt{True} \vee \mathtt{False}, \mathtt{False}) |} \cdot 100.
\end{align}
The lower the metric, the less ignorant the model is. Additionally, we combine TU and ID using a harmonic mean to make it easier to compare on our benchmark.
We refer to it as F1-Score, which is computed as follows: $F1=\frac{2}{TU^{-1} + (100-ID)^{-1}}$. 

\section{Experiment}\label{sec:exp}
In \Sref{sec:main_result}, we evaluate the current VLMs using our BEAF dataset. We employ both proposed and traditional metrics and compare their performance with other benchmarks. 
Furthermore, we conduct an analysis of image-wise dynamic correctness by visualizing image-wise object relationship containing vision and text axes. We provide a report on the object-wise error rate.
In \Sref{sec:add_exp}, we assess the performance of VLMs on the open generation task using our dataset, while also analyzing both the dataset and VLMs based on the CLIP score. 

All the reported performance is the zero-shot performance of VLMs, including LLaVA~\cite{liu2023improvedllava}, InstructBLIP~\cite{instructblip}, Shikra~\cite{chen2023shikra}, and mPLUG-Owl~\cite{ye2023mplug}.
We use prompt templates provided in their released codebases or papers.

\subsection{Main Results}\label{sec:main_result}

\paragraph{Evaluation on BEAF benchmark}
We evaluate several VLMs on our BEAF dataset using the proposed change-aware metrics, TU, IG, SB, ID, and F1.
The evaluation results are presented in \Tref{tab:main}. Among the 13B-size models, LLaVA outperforms InstructBLIP in terms of TU, IG, and SB because InstructBLIP consistently answers ``\texttt{Yes}'' regardless of scene manipulation, and this result is related to its high value of SB$_p$. InstructBLIP achieves better ID compared to LLaVA. Among 7B-size models, Shikra exhibits the best in TU and SB$_{p}$. Shikra is trained using a location-aware strategy, which could be beneficial for hallucination evaluation because knowledge of location aids in judging the existence of an object. The overall results indicate that, although it is rare to get all the answers wrong, it is common for the answers to remain the same even though the image has changed. This indicates that recent VLMs are not adequately reflective of such changes.

\begin{table}[t]
    \centering
    \caption{\textbf{Evaluation results on BEAF benchmark.} 
    We evaluate various VLMs on our benchmark.
    F1 is a harmonic mean of TU and 1-ID.
    Yes ratio is the percentage of total answers that are answered ``\texttt{Yes}'' for the removed object-related question. The former is for the original image, and the latter stands for the manipulated image.
    \underline{Underline} stands for the best.}
    \label{tab:main}
    \resizebox{1.0\textwidth}{!}{ 
            \begin{tabular}{c@{\quad}l@{\quad\quad}C{16mm}C{16mm}C{16mm}C{16mm}C{16mm}C{16mm}C{20mm}}
            \toprule
            \textbf{Size} & \textbf{Model} & \textbf{TU($\uparrow$)} & \textbf{IG($\downarrow$)} & \textbf{SB}$_{p}$($\downarrow$)& \textbf{SB}$_{n}$($\downarrow$)& \textbf{ID}($\downarrow$) & \textbf{F1($\uparrow$)} &\textbf{Yes (\%)}\\ \midrule
            \multirow{2}{*}{13B}
            & LLaVA-v1.5   & \underline{24.3} & \underline{0.2} & \underline{72.0} & \underline{3.5} & 6.4 & \underline{38.6} & 96.4 $\rightarrow$ 72.2\\
            & InstructBLIP & 11.1 & 0.3 & 84.4 & 4.2 & \underline{6.1} & 19.9 & 95.5 $\rightarrow$ 84.7 \\
            \midrule
            \multirow{4}{*}{7B}
            & LLaVA-v1.5   & 32.6 & 0.1 & 61.3 & 6.0 & 5.6 & 48.5 & 93.9 $\rightarrow$ 61.4 \\
            & InstructBLIP & 33.5 & 0.6 & 51.3 & 14.6 & \underline{5.2} & 49.5 & 84.8 $\rightarrow$ 51.9\\
            & Shikra       & \underline{52.7} & 0.4 & \underline{31.2} & 15.7 & 5.7 & \underline{67.6} & 83.9 $\rightarrow$ 31.6 \\
            & mPLUG-Owl2   & 24.6 & \underline{0.0} & 72.1 & \underline{3.3} & 7.0 & 38.8 & 96.7 $\rightarrow$ 72.1 \\
            \bottomrule
            \end{tabular}
            }
\end{table}

\begin{table}[t]
    \centering
    \caption{\textbf{Traditional evaluation results on BEAF dataset.} 
    We evaluate various VLMs on our dataset in terms of accuracy, precision, recall, and F1 score.
    Note that the BEAF dataset includes both original and manipulated data but does not consider the changes for this traditional evaluation.
    \underline{Underline} stands for the best.}
    \label{tab:tra_main}
    \resizebox{1.0\textwidth}{!}{ 
            \begin{tabular}{c@{\quad}l@{\quad}C{20mm}C{20mm}C{20mm}C{20mm}C{20mm}}
            \toprule
            \textbf{Size} & \textbf{Model} & \textbf{Acc.($\uparrow$)} & \textbf{Precision($\uparrow$)} & \textbf{Recall($\uparrow$)} & \textbf{F1($\uparrow$)} & \textbf{Yes (\%)} \\
            \midrule
            \multirow{2}{*}{13B}
            & LLaVA-v1.5   & \underline{74.9} & \underline{56.3} & 93.4 &\underline{70.3} & 52.6 \\
            & InstructBLIP & 71.5 & 52.8 & \underline{94.3} & 67.7 & 56.7 \\
            \midrule
            \multirow{4}{*}{7B}
            & LLaVA-v1.5   & 79.7 & 62.6 & 89.5 & 73.7  & 45.4 \\
            & InstructBLIP & 82.0 & 68.3 & 80.7 & 74.0 & 37.5 \\
            & Shikra       & \underline{84.5} & \underline{74.9} & 76.9 & \underline{75.9} & 32.6 \\
            & mPLUG-Owl2   & 69.1 & 50.7 & \underline{93.8} & 65.8 & 58.7 \\
            \bottomrule
            \end{tabular}
            }
\end{table}

\begin{figure}[t]
    \centering
    \includegraphics[width=1\linewidth]{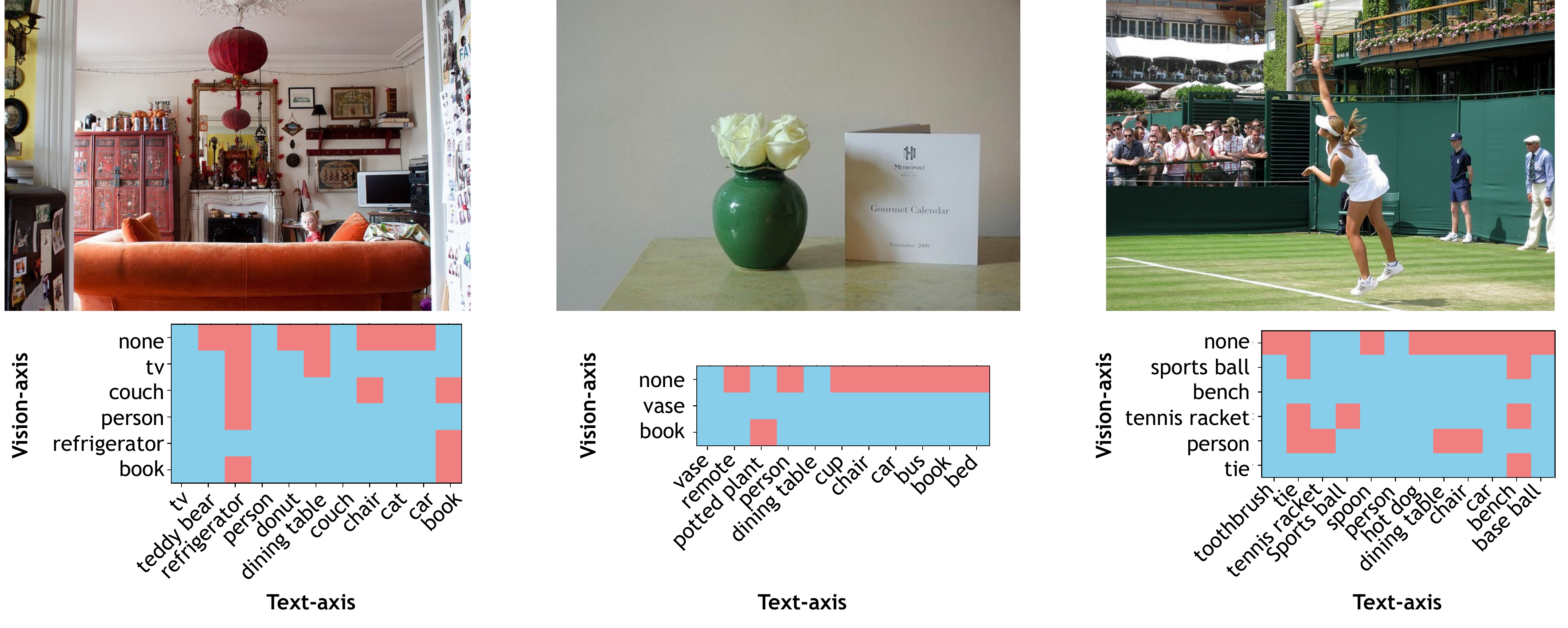}
    \caption{\textbf{Visualization of image-wise object relationship.} 
    We visualize image-wise object relationship along the text and vision axes from Shikra.
    [Top] Original image samples. [Bottom] The object relation table given the manipulated images and questions.
    We color the correct answer blue and the wrong one red.
    The text axis stands for the target object queried in the question, and the vision axis for the removed object in the image.
    The ``none'' in the vision axis means the original image (not manipulated).
    Thereby, we can analyze influence between objects within a scene at once.
    }
    \label{fig:visualiation}
\end{figure}

\begin{figure}[t]
    \centering
    \includegraphics[width=0.83\linewidth]{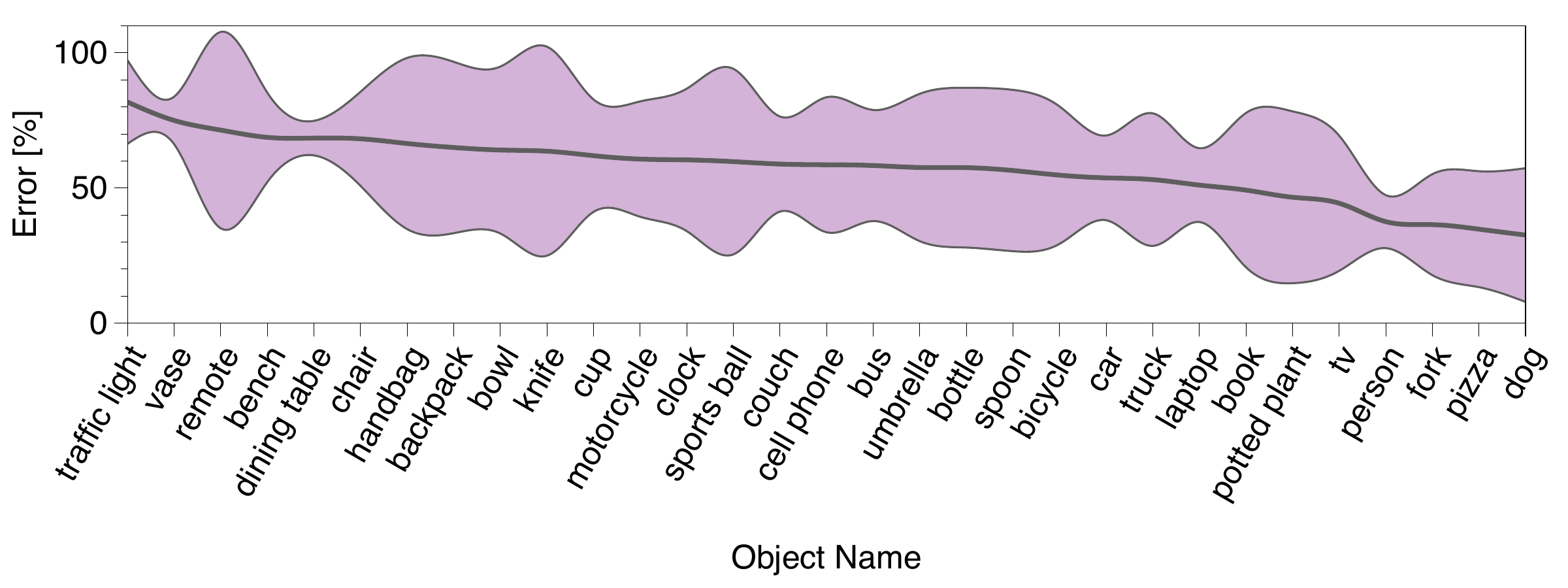}
    \caption{\textbf{Object-wise error rate on the manipulated image-question pairs.}
    We plot the error rate of each object to investigate which object is inaccurately inferred when it is removed.
    We exclude the objects manipulated less than 20 times.
    The solid line represents the average error rate of VLMs (7B), while the purple area indicates the 95\% confidence interval.
    }
    \label{fig:err_graph}
\end{figure}

\paragraph{Traditional evaluation on BEAF dataset}
In addition to the proposed change-aware metrics, we also evaluate using traditional metrics such as accuracy, precision, recall, and F1 score, as shown in \Tref{tab:tra_main}.
While the trends in accuracy or F1 score are similar to those shown in TU results of \Tref{tab:main}, the absolute values of TU are much lower than those of the traditional metrics.
This discrepancy suggests that, although a model achieves high performance in 
the existing benchmarks, our benchmark tells us that we cannot conclude low hallucination and the models may still suffer from hallucination.
For example, although LLaVA-v1.5-13B achieves high performance on accuracy, the Yes ratio in \Tref{tab:main} remains high even after the manipulation.
It may include instances where the correct answer was guessed without actual reference to the given image.
The experimental findings underscore the importance of observing changes in answers along the vision-axis to accurately evaluate hallucination in VLMs.

\setlength{\intextsep}{0pt}%
\begin{wraptable}{r}{0.5\linewidth}
    \centering
    \caption{\textbf{Comparison of performance on VQA and TU.} 
    We compare the accuracy on VQA datasets, VQAv2 and GQA, as well as TU on BEAF dataset. 
    }
    \label{tab:other_benchmark}
    \resizebox{0.85\linewidth}{!}{ 
    \begin{tabular}{@{\ \ }l@{\ \ \ }C{13mm}C{12mm}C{11mm}}
        \toprule
        \textbf{Model (7B)} & \textbf{VQAv2} & \textbf{GQA} & \textbf{TU} \\
        \midrule
        LLaVA-v1.5 & 78.5 & 62.0 & 32.6\\
        InstructBLIP & - & 49.2 & 33.5\\
        Shikra & 77.4 & - & 52.7 \\
        mPLUG-Owl2 & 79.4 & 56.1 & 24.6\\
        \bottomrule
    \end{tabular}}
\end{wraptable}
\paragraph{Comparison with other benchmarks} 
We compare our metric TU with the performance for other benchmarks of VQAv2~\cite{balanced_vqa_v2} and GQA~\cite{gqa2019}, as shown in \Tref{tab:other_benchmark}.
Each VQA performance value is reported from the proposed papers.
The results indicate that, while mPLUG-Owl2 outperforms others on VQAv2 and LLaVA excels on GQA, Shikra leads in performance on our metric.
We observe that there is a negative correlation between performance and the incidence of hallucination, which aligns with
the previous observation~\cite{wang2023evaluation}.
Although improving performance on various benchmarks is challenging, it is essential to check multiple aspects of the performance, including hallucination.
We believe that our benchmark provides valuable insights in conjunction with existing benchmarks.

\paragraph{Image-wise object relationship}
We visualize changes in dynamic object relationship along the vision and text axes, as shown in \Fref{fig:visualiation}. 
Compared to our metrics providing comprehensive statistics as summarized scores, this visualization enables to track the changes in VLMs' responses at object-wise and scene-wise levels.
The x-axis is the text axis where each object is associated with the question, ``Is there \{object\}?''; the y-axis is the vision axis where the object is removed from the original image. 
The blue color is correct, the red one is wrong.

In the first column in \Fref{fig:visualiation}, altering the scene affects the textual correctness. Specifically, removing either the couch or the book shifts the book's textual correctness from correct to incorrect.  This suggests a correlation between the couch and the book, indicating that complex scenes can lead to model hallucinations even in the absence of the book. In the second scene, when the object vase or book is removed, most of the incorrectness becomes correct. This could imply that the presence of a vase or a book might lead the model to misinterpret the scene. Through the simultaneous analysis of both vision and text axes, we observe immediate changes in the answers and offer interpretations for the dynamic interactions observed in the second and third scenes illustrated in the figure.

\paragraph{Object-wise error rate} 
We compute the mean and 95\% confidence interval of the object-wise error after the manipulation in \Fref{fig:err_graph}. 
It represents whether the VLMs can accurately detect the removal.
The results show that VLMs frequently fail to recognize certain objects like traffic lights, vases, and remotes.
We speculate that the difficulty in recognizing these objects may stem from their complexity or their infrequent appearance in the training data.

\begin{figure}[t]
    \centering
    \includegraphics[width=1.0\linewidth]{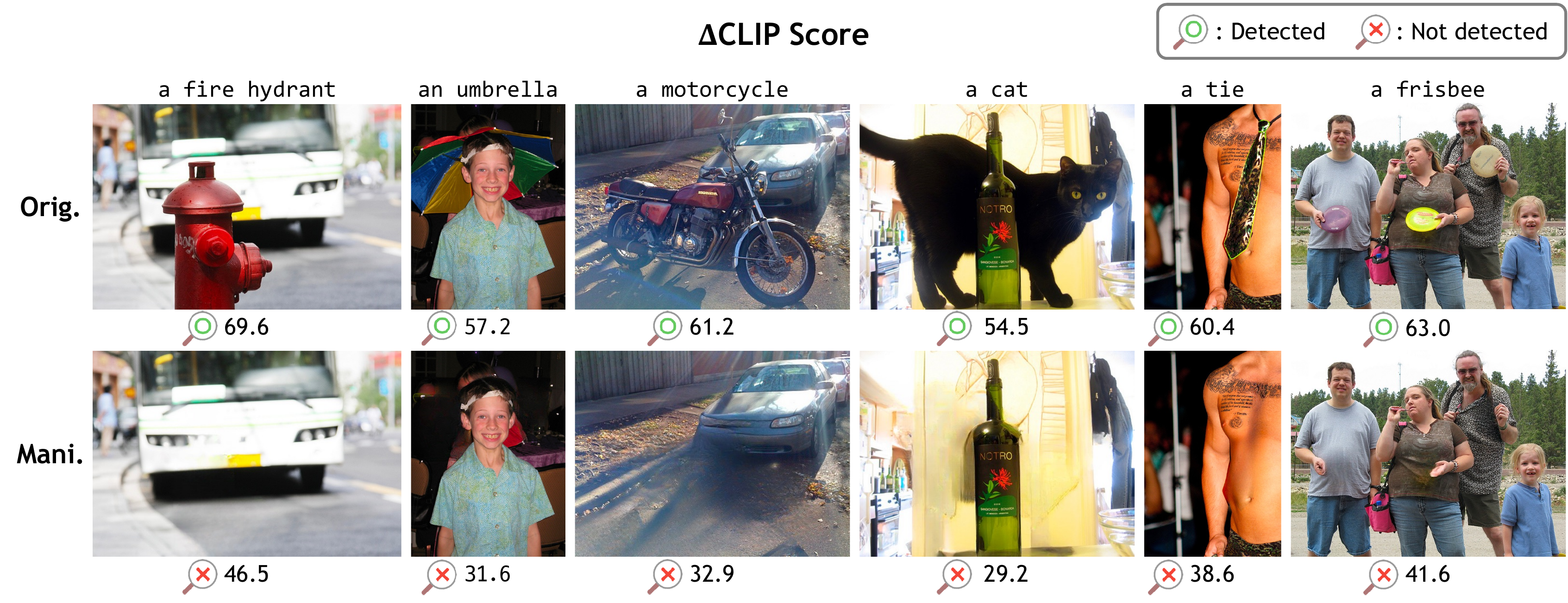}
    \caption{\textbf{Samples with CLIP Score.}
    We visualize samples and display the CLIP Score below.
    Additionally, we report the results of Faster R-CNN~\cite{ren2015faster} using the magnifying glass icon.
    After removing the object, the removed object is not detected, which demonstrates the manipulation quality.
    }
    \label{fig:clipscore_ex}
\end{figure}

\begin{figure}[t]
    \centering
    \includegraphics[width=0.9\linewidth]{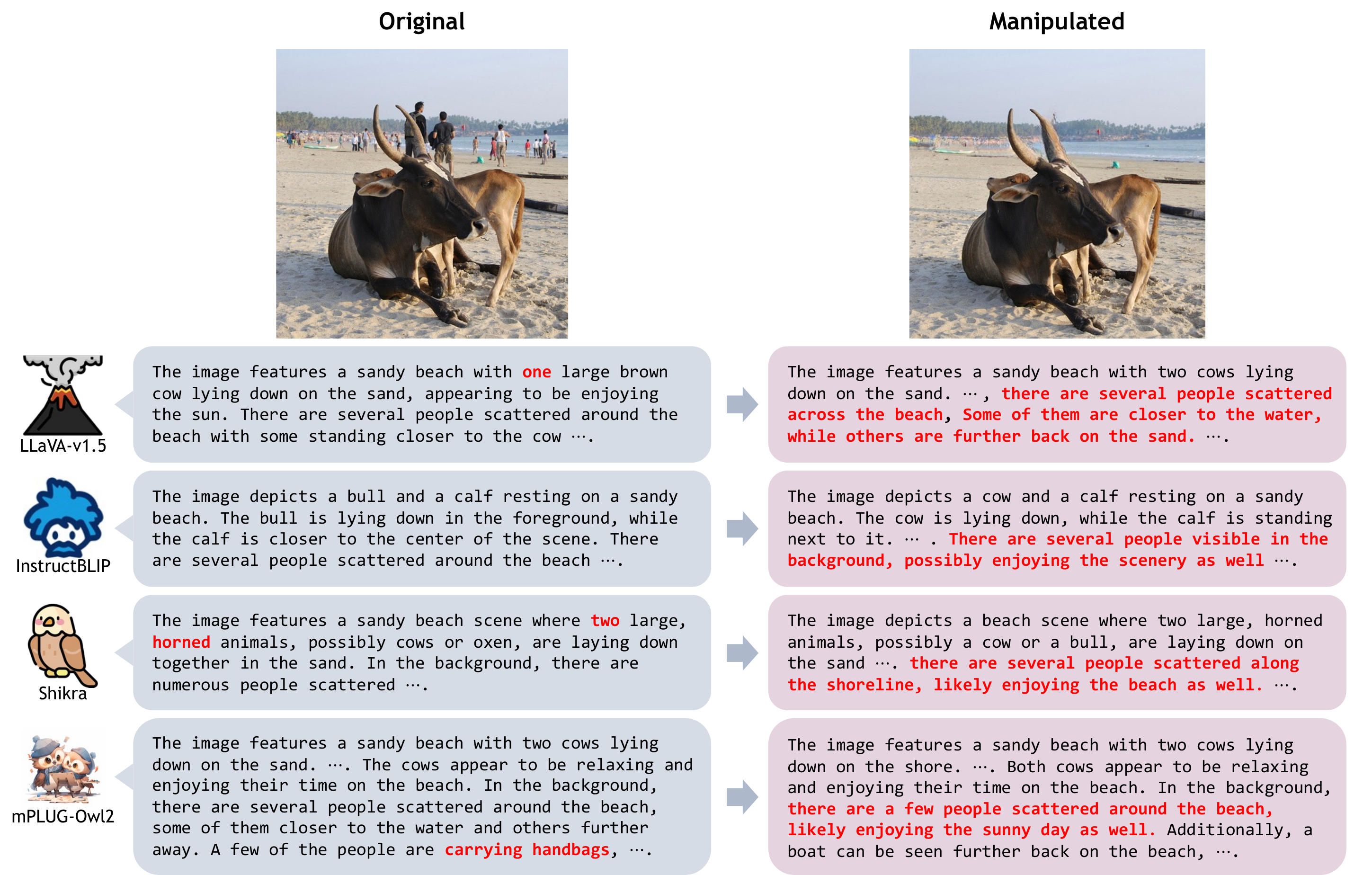}
    \caption{\textbf{Qualitative example on open-ended generation.} 
    We report the changes in answers for the open-ended generation task with the prompt ``\texttt{Describe this image.}'' as the image undergoes manipulation.
    The red color highlights the hallucination parts.
    } 
    \label{fig:qual}
    
\end{figure}

\subsection{Additional Study}\label{sec:add_exp}
\

\begin{wraptable}{r}{0.45\linewidth}
    \centering
    \caption{\textbf{Comparison of CHAIR metric and TU.} 
    We measure the CHAIR score based on the answers derived from the images in BEAF using the prompt ``\texttt{Describe this image.}''}
    \label{tab:chair}
    \resizebox{1.0\linewidth}{!}{ 
        \begin{tabular}{c@{\ \ }lC{16mm}C{16mm}C{14mm}}
        \toprule
        \textbf{Size} & \textbf{Model} & \textbf{CHAIR$_I$} & \textbf{CHAIR$_S$} & \textbf{TU}\\ \midrule
        \multirow{2}{*}{13B}
        & LLaVA-v1.5 & 21.6 & 72.3 & 24.3 \\
        & InstructBLIP & 28.1 & 81.1 & 11.1\\
        \midrule
        \multirow{4}{*}{7B}
        & LLaVA-v1.5 & 22.3 & 73.9 & 32.6\\
        & InstructBLIP & 21.0 & 67.5 & 33.5 \\
        & Shikra & 23.6 & 77.0 & 52.7 \\
        & mPLUG-Owl2 & 25.1 & 78.4 & 24.6 \\
        \bottomrule
        \end{tabular}
    }
\end{wraptable}
\paragraph{Evaluation of open generation} 
In \Tref{tab:chair}, we compare the results of the CHAIR~\cite{rohrbach2018chair} and TU metrics to evaluate the hallucination in the open-ended generation answers on our dataset.
Briefly, CHAIR$_I$ and CHAIR$_S$ count the hallucinated instances and sentences containing these objects in the generated output, respectively.
The lower the score, the better.
We observe that if the open generation contains the hallucinated object and sentences, TU is also lower, as seen with InstructBLIP-13B.

\begin{wraptable}{r}{0.44\linewidth}
    \centering
    \caption{\textbf{CLIP Score.} We measure the CLIP Score of original and manipulated images with the text prompt.}\label{tab:clipscore}
    \resizebox{0.8\linewidth}{!}{ 
        \begin{tabular}{l@{\quad}C{24mm}}
        \toprule
        \textbf{Type} & \textbf{CLIP Score} \\ \midrule
        Original & 53.7 \\
        Manipulated & 47.9 \\ \bottomrule
        \end{tabular}
    }
\end{wraptable}
\noindent\textbf{CILP Score.} 
We compare the CLIP Score \cite{hessel2021clipscore} between original and manipulated images to ensure the target object has been well-removed.
A lot of VLMs utilize the visual encoder from CLIP, prompting us to analyze CLIP's performance with our datasets.
We measure the CLIP score between an object and a prompt ``\texttt{a photo of \{object\}.}''
We use a cropped object image based on the GT bounding box. 
If the image has multiple objects of the same semantic class, we use the average CLIP score of all the same semantic objects.
For the manipulated image, \texttt{\{object\}} is the removed object.
As shown in \Tref{tab:clipscore}, the manipulated images have lower CLIP scores compared to their original counterparts.

We also visualize the images with the CLIP Score and the multi-class classification results in \Fref{fig:clipscore_ex}.
The multi-class classification correctly identifies the depicted objects.
In addition, \Fref{fig:qual} visualizes an open generation sample result, where red-colored text indicates hallucinated words.
Before manipulation, all the VLMs mention ``people'' in the background; however, they still mention ``people'' even though all the persons are removed from the image. This implies that the commonly used visual encoder might be a non-negligible source of hallucination.

\section{Discussion and Conclusion}
We present a comprehensive analysis of hallucination using our BEAF benchmark.
Unlike conventional hallucination benchmarks that measure accuracy by changing only the text-axis question, we also manipulate images by removing objects and propose four sophisticated metrics to track the changes in answers according to image changes.
This approach enables us to a more discerning analysis of the hallucination in two-axis views, $\textit{e.g.}$, vision and text axes.
We can find with our metric that traditional accuracy may include instances where the correct answer was guessed without actual reference to the given image.
We found that our TU is a more \emph{calibrated} metric, which allows us to observe the intringuing phenomenon that has not been revealed in the prior work: 
VLMs tend to be confused if the complexity of an image is high.
(refer to Sec. \textcolor{red}{A} in the supplementary material).
Additionally, we introduce the visualization method to track the correctness as the objects are removed.
We believe that our benchmark presents a comprehensive analysis.

\paragraph{Limitation}
While our benchmark offers a more fine-grained evaluation,
it still faces the constraints of the MS-COCO dataset. These include a limited variety of object categories and an uneven distribution of object instances. 
While we utilized current state-of-the-art in-painting models and mask extractors, we manually adjusted the images to ensure image quality and dataset integrity.
This step prevented us from achieving a fully automated creation process.
Although such limitations exist, these aspects are expected to naturally diminish or disappear as in-painting models advance, making our dataset construction pipeline fully automatic and our method applicable to all images.  Furthermore, we foresee future evaluations expanding in scale through manipulations like object removal, insertion, or resizing.  While our current method may appear as a modest evaluation benchmark, we hope that our work becomes the first step toward controllable hallucination evaluation.

\section*{Acknowledgment}
This work was supported by Institute of Information \& communications Technology Planning \& Evaluation (IITP) grant funded by the Korea government(MSIT) (No.2021-0-02068, Artificial Intelligence Innovation Hub; No.2022-0-00124, Development of Artificial Intelligence Technology for Self-Improving Competency-Aware Learning Capabilities; No.RS-2019-II191906, Artificial Intelligence Graduate School Program(POSTECH))

%
%
\bibliographystyle{splncs04}
\bibliography{main}
\end{document}